\documentclass[runningheads]{llncs}
\usepackage{times}
\usepackage{latexsym}
\usepackage{bm}
\usepackage{amsmath}
\usepackage{amsfonts}
\usepackage{booktabs}
\usepackage{array}
\usepackage{multirow}
\usepackage{graphicx}
\usepackage{color}
\usepackage{subfigure}
\usepackage{url}
\usepackage{hyperref}
\usepackage{breakurl}
\usepackage{CJKutf8}

\begin{document}
\begin{CJK}{UTF8}{gbsn}	
%
\title{CCKS 2019 Shared Task on Inter-Personal Relationship Extraction}
\titlerunning{CCKS 2019 Shared Task on IPRE}
\author{Haitao Wang\inst{1}, Zhengqiu He\inst{1}, Tong Zhu\inst{1}, Hao Shao\inst{2}, Wenliang Chen\inst{1}, Min Zhang\inst{1}}
\authorrunning{H. Wang et al.}
\institute{School of Computer Science and Technology, Soochow University, China \\
	\email{\{htwang2019,zqhe,tzhu7\}@stu.suda.edu.cn,\\
		\{wlchen, minzhang\}@suda.edu.cn}
	\and Gowild Robotics Co., Ltd, China \\
    \email{shaohao820@gmail.com}}
%
%

\maketitle              

\begin{abstract}
The CCKS2019 shared task was devoted to  inter-personal relationship extraction. 
Given two person entities and at least one sentence containing these two entities, participating teams are asked to predict the relationship between the entities according to a given relation list. 
This year, 358 teams from various universities and organizations participated in this task. 
In this paper, we present the task definition, the description of data and the evaluation methodology used during this shared task. 
We also present a brief overview of the various methods adopted by the participating teams. 
Finally, we present the evaluation results.

\end{abstract}
\section{Introduction}
\label{section_introduction}

Inter-personal relationship extraction is one of the shared task of China Conference on Knowledge Graph and Semantic Computing in 2019 (CCKS2019).
The goal of this task is to predict the relationship between two given person entities from sentences, which contain these two entities.

This task is an attractive and potential research direction. On one hand, as an important sub-task of information extraction, relation extraction has always been an important basis for many intelligent applications, such as QA and search. On the other hand, as the basis of society as a whole, inter-personal relationship is a strong, deep or close connection among persons \cite{white2016evolution}.

At present, relation extraction with supervised paradigm \cite{feng2018reinforcement,he2019syntax,ji2017distant,zhou2016attention} has been widely used and yielded relatively high performance.
However, supervised relation extraction systems require a masssive amount of labeled training data, which is time-consuming and laborious. 
To solve this problem, Mintz et.al \cite{mintz2009distant} proposed the concept of distant supervision.
They assumed that if two entities $e_h$ and $e_t$ have a relation $r$ in the knowledge base such as Freebase \cite{bollacker2008freebase}, then all sentences mentioning these two entities will express the relation $r$ in some way.
Through distant supervison, we can automatically generate large amounts of labeled data via aligning knowledge bases and texts.

However, it is inevitable that data generated by supervison has wrong labeling problems due to strong distant supervision assumption, which may lead to wrong evaluations.
In order to address the problem, we can hire workers to label the test data.
Then we use a dataset, named as Inter-Personal Relationship Extraction (IPRE) \cite{wang2019ipre} in Chinese, in which
manually annotated data is used as development and test sets.
The CCKS2019 shared task provides a platform for participating teams to evaluate the methods of inter-personal relation extraction, and the data is available at \url{https://github.com/SUDA-HLT/IPRE}.

The rest of this paper is organized as follows.
In Section \ref{section_task}, we provide the definition of task.
In Section \ref{section_data}, we describe how data is prepared, including the format of data and the process of data construction and annotation.
We also introduce the evaluation methodology in Section \ref{section_evaluation}.
In Section \ref{section_approaches},we give an introduction to the approaches used by participating teams and shows the ranking of top 10.
Finally, we present our conclusions in Section \ref{section_conclusion}.


\section{Task Definition}
\label{section_task}

The shared task on inter-personal relationship extraction is to identify the relationship between people. 
Specifically, given one or more sentences and a list of inter-personal relationships, a relation extraction system is asked to identify the relationship between two designated entities from the sentences.

Since the data is generated by distant supervision, we can predict the relation between person entities at bag-level. 
Here, all sentences that mention the same entity pair are regarded as a bag. 
Nevertheless, we can also predict relation at sentence-level when testing because development and test sets are manually labeled.
Hence, the shared task consists of two sub-task: sent-track and bag-track.

\subsection{Sent-Track}
In the sent-track, all participating teams are asked to predict the relations at sentence-level in the test data. For instance, in the following Example \ref{exam1}, given a sentence and an entity pair $<$贾玲, 冯巩$>$, the relation between ``贾玲" (Jia Ling) and ``冯巩" (Feng Gong) can be detected as \emph{teacher} (老师).
\begin{example}
	\label{exam1}
	\hspace*{\fill} \\
	\textbf{Input}\quad$<$贾玲, 冯巩$>$\\
	\setlength{\parindent}{5em}\quad \textbf{贾玲}，80后相声新秀，师承中国著名相声表演艺术家\textbf{冯巩}。\\
	\textbf{Output}\quad\emph{teacher} (老师)
\end{example}

\subsection{Bag-Track}
In the bag-track, each bag contains at least one sentence and relation prediction is based on bags. 
Such as Example \ref{exam2} , given an entity pair $<$袁汤, 袁安$>$ and several sentences contains ``袁汤" (Yuan Tang) and ``袁安" (Yuan An). 
The correct relation beween these two person entities is \emph{paternal grandfather} (爷爷).
Sentence 2 can express this relation while the sentence 1 cannot.
The relation expressed by sentences in a bag is considered as the label of the whole bag.
Both sent-track and bag-track require contestants to predict relations as much as possible.
\begin{example}
	\label{exam2}
	\hspace*{\fill} \\
	\textbf{Input}\quad$<$袁汤, 袁安$>$\\
	1. 从\textbf{袁安}起，几代位列三公(司徒、司空、太尉)，出过诸如\textbf{袁汤}、袁绍、袁术等历史上著名人物。\\ 
	2. \textbf{袁汤}（公元67年—153年），字仲河，河南汝阳（今河南商水西南人，名臣\textbf{袁安}之孙，其家族为东汉时期的汝南袁氏。\\
	\textbf{Output}\quad\emph{paternal grandfather} (爷爷)
\end{example}
\section{Data}
\label{section_data}

In this section, we briefly introduce the  data used in this shared task and the detailed information can be found in \cite{wang2019ipre}.

\subsection{Data Format}
Different from the format of some datasets widely used in relation extraction research, such as Riedel2010 \cite{riedel2010modeling}, ACE05 and CONLL04 \cite{roth2004linear}, in order to make the contestants have a clearer and orderly understanding of the data, we have made some abstract representations of the datasets. 
Each relation is given a \emph{relationID} and each instance in each dataset is given a unique \emph{sentID}. 
Each bag grouped by the sentences with the same entity pair is also given a \emph{bagID}.
Here, each instance consists of one ordered entity pair  $<\!\!e_h, e_t\!\!>$ and one sentence.
Sentences in all datasets are segmented, and the result of sentence segmentation is for reference only, contestants may use it as appropriate.
Table \ref{data_format} shows the file format infomation in dataset.
Each part mentioned above is separated by a tab character.
Moreover, \emph{relationIDs} is composed of several \emph{relationID} that separated by a space character, and \emph{sentIDs} is also so.
\begin{table}[tb]
	\centering
	\footnotesize
	\caption{Format of data set files.}
	\label{data_format}
	\begin{tabular}{|l|l|}
		\hline
		\multicolumn{1}{|c|}{\textbf{Files}} & \multicolumn{1}{c|}{\textbf{Contents}} \\ \hline
		sent\_train/dev        & sentID\ \ $e_h$\ \ $e_t$\ \ sentence                      \\
		sent\_relation\_train/dev &  sentID\ \ relationIDs \\
		bag\_relation\_train/dev &  bagID\ \ $e_h$\ \ $e_t$\ \ sentIDs\ \ relationIDs       \\ \hline
		sent\_test          & sentID\ \ $e_h$\ \ $e_t$\ \ sentence                       \\
		sent\_relation\_test  & sentID      \\
		bag\_relation\_test &  bagID\ \ $e_h$\ \ $e_t$\ \ sentIDs                    \\ \hline
	\end{tabular}%
\end{table}

In addition to the dataset, we also provide a large-scale text file consisting of 10 million unlabeled sentences, which can be used to train language models or word vectors.
Except for the data mentioned above, contestants cannot use other annotated data.

\begin{table}[tph]
	\centering
	\footnotesize
	\caption{Relation Types in IPR}
	\label{relations}
	\begin{tabular}{|c|l|c|c|c|c|c|c|}
		\hline
		\multirow{2}{*}{RelationID} & \multicolumn{1}{c|}{\multirow{2}{*}{Relation type}} & \multicolumn{2}{c|}{Train} & \multicolumn{2}{c|}{Dev} & \multicolumn{2}{c|}{Test} \\ \cline{3-8} 
		& \multicolumn{1}{c|}{} &  Sentence  & Triple &  Sentence  &\ Triple  &  Sentence  &  Triple \\ \hline
		\multirow{2}{*}{\# 0}
		& NA (distant superversion) & 248,850 & 35,000 & 35,550 & 5,000 & 71,091 & 9,999 \\ \cline{2-8} 
		& NA (human annotation) & 0 & 0 & 1782 & 354 & 3683 & 726 \\ \hline
		\# 1 & 现夫 (present husband) & 8,142 & 407 & 218 & 52 & 536 & 112 \\ \hline
		\# 2 & 前夫 (former husband) & 218 & 20 & 11 & 4 & 12 & 7 \\ \hline
		\# 3 & 未婚夫 (betrothed husband) & 183 & 10 & 5 & 4 & 13 & 11 \\ \hline
		\# 4 & 现妻 (present wife) & 5,544 & 260 & 125 & 32 & 296 & 71 \\ \hline
		\# 5 & 前妻 (former wife) & 245 & 13 & 3 & 3 & 3 & 3 \\ \hline
		\# 6 & 未婚妻 (betrothed wife) & 69 & 7 & 16 & 4 & 10 & 9 \\ \hline
		\# 7 & 爷爷 (paternal grandfather) & 291 & 60 & 9 & 6 & 9 & 7 \\ \hline
		\# 8 & 奶奶 (paternal grandmother)& 40 & 4 & 1 & 1 & 1 & 1 \\ \hline
		\# 9 & 外公 (maternal grandfather) & 9 & 4 & 3 & 1 & 0 & 0 \\ \hline
		\# 10 & 生父 (father) & 6,870 & 759 & 247 & 73 & 490 & 154 \\ \hline
		\# 11 & 生母 (mother) & 1,383 & 169 & 65 & 18 & 107 & 39 \\ \hline
		\# 12 & 儿子 (son) & 2,627 & 217 & 88 & 26 & 221 & 51 \\ \hline
		\# 13 & 女儿 (daughter) & 830 & 105 & 43 & 7 & 63 & 28 \\ \hline
		\# 14 & 孙子 (paternal grandson) & 46 & 12 & 4 & 2 & 4 & 2 \\ \hline
		\# 15 & 孙女 (paternal granddaughter) & 19 & 5 & 2 & 1 & 8 & 2 \\ \hline
		\# 16 & 哥哥 (elder brother) & 1,673 & 142 & 27 & 9 & 43 & 24 \\ \hline
		\# 17 & 弟弟 (younger brother) & 637 & 77 & 11 & 6 & 39 & 14 \\ \hline
		\# 18 & 姐姐 (elder sister) & 532 & 53 & 4 & 3 & 9 & 6 \\ \hline
		\# 19 & 妹妹 (younger sister) & 805 & 69 & 9 & 6 & 23 & 11 \\ \hline
		\# 20 & 叔伯 (father's brother) & 77 & 8 & 1 & 1 & 1 & 1 \\ \hline
		\# 21 & 舅舅 (mother's brother) & 77 & 5 & 1 & 1 & 0 & 0 \\ \hline
		\# 22 & 姑妈 (father's sister) & 22 & 5 & 11 & 1 & 2 & 2 \\ \hline
		\# 23 & 侄子 (brother's son) & 158 & 9 & 1 & 1 & 0 & 0 \\ \hline
		\# 24 & 侄女 (brother's daughter) & 30 & 6 & 2 & 1 & 0 & 0 \\ \hline
		\# 25 & 儿媳 (daughter-in-law) & 13 & 7 & 1 & 1 & 0 & 0 \\ \hline
		\# 26 & 女婿 (son-in-law) & 119 & 8 & 1 & 1 & 4 & 2 \\ \hline
		\# 27 & 嫂子 (elder brother's wife) & 67 & 8 & 1 & 1 & 0 & 0 \\ \hline
		\# 28 & 公公 (husband's father) & 24 & 6 & 1 & 1 & 0 & 0 \\ \hline
		\# 29 & 岳父 (wife's father) & 165 & 12 & 2 & 1 & 1 & 1 \\ \hline
		\# 30 & 朋友 (friend) & 1,610 & 71 & 6 & 5 & 27 & 12 \\ \hline
		\# 31 & 喜欢 (love) & 1,301 & 78 & 27 & 16 & 44 & 29 \\ \hline
		\# 32 & 恋人 (lover) & 1,266 & 94 & 72 & 37 & 149 & 72 \\ \hline
		\# 33 & 老师 (teacher) & 2,911 & 206 & 62 & 20 & 201 & 61 \\ \hline
		\# 34 & 学生 (student) & 547 & 35 & 5 & 3 & 11 & 8 \\ \hline
	\end{tabular}
\end{table}

\subsection{IPRE Dataset}
The data of the shared task mainly comes from the Internet webpage text, in which the develpment set and the test set are manually annotated, and the training set is automatically generated by distant supervision.

\subsubsection{Data Construction}
Since there is no Chinese knowledge base like Freebase that provides enough inter-personal relation triples for text alignment, we need to extract triples by ourselves, and wiki-style website (e.g., Wikipedia and Chinese Baidu Baike) can satisfy the demands of data construction.
We screen all the entities in the webpages and construct a high-quality person entity list by multiple information verification.
Then, we construct a list with 942,344 person entity in all.
According to the person entity list, we select potential triples from infobox of the webpages, and both entities appeared in the triples are in the person entity list. 
We maually merge and denoise the expressions of relations that exist in these triples, and define a set of inter-personal relations (IPR), as shown in Table \ref{relations}.
After constructing the person entity list and extracting the triples, we align the triples to the webpages. 
Finally, we obtain over 41,000 sentences and 4,214 bags.

\begin{table}[tb]
	\centering
	\footnotesize
	\caption{Dataset partition and annotation in IPRE. 
	}
	\label{data_pa}
	\begin{tabular}{|c|c|c|c|c|}
		\hline
		\ \ Type\ \  & \ \ Percentage\ \  & \ \  Bag \ \ &\ \  Distant Supervision\ \  &\ \  Human Annotation\ \  \\
		\hline
		train & 70\% &2948 & \checkmark & $\times$  \\
		\hline
		dev & 10\%   &416 & \checkmark & \checkmark  \\
		\hline
		test & 20\%  &850 & \checkmark & \checkmark  \\
		\hline
	\end{tabular}
\end{table}

\subsubsection{Data Annotation}
After obtaining the data generated by the distant supervision, we divide this dataset into a training set (70\%), a development set (10\%) and a test set (20\%), ensuring that there is no overlap in any two of them.
As discribled in Section \ref{section_introduction}, we manually annotate all the sentences in development set and test set.

9,776 sentences are manually annotated, in which 899 sentences are labeled as illegal sentences.
In the remaining sentences, 3,412 sentences express at least one of 34 relation types, while the others are labeled as NA.
At bag level, these legal sentences are divided into 1,266 bags, of which 941 bags express valid relation types. 
The detailed information is listed in Table \ref{data_pa}.

\section{Evaluation}
\label{section_evaluation}
This section introduces the evaluation metrics used in the shared task and how to rank according to scores. 

\subsection{Evaluation Metrics}

Since non-NA data in the test set is labeled manually, we can calculate the F1-score of the predictions submitted by the participating teams. 
In this shared task on inter-personal relationship extraction, teams are ranked by F1-score both in sent-track and bag-track.
When calculating the F1-score, prediction results of NA data are not considered.
Or, more accurately, if a sentence or a bag is predicted to have more than one relation label, whether it has a NA label or not will not affect the calcultion of F1-score.
However, once the NA relation is predicted to be a non-NA relation, the F1-score will decrease due to the decline in precision of relation extraction system.

\subsubsection{Evaluation for Sent-Track}
In sent-track, given a prediction result at sentence-level on test data, the number of sentences that are predicted to be non-NA results is recorded as $N_{sys}$ without regard to NA, in which the number of correctly predicted sentences is recorded as $N_r$.
If there are multiple non-NA relations in the prediction result of one sentence, such as two non-NA relations, then it will be counted as twice when calculating $N_{sys}$.
The number of sentences in the standard answer is $N_{std}$, which is a constant.
Thus, the evaluation metrics of sent-track are defined as:
\begin{equation}
\label{evalution_metrics}
\begin{split}
P   = \frac{N_r}{N_{sys}}, 
R   = \frac{N_r}{N_{std}}, 
F_1   = \frac{2PR}{P+R}
\end{split}
\end{equation}

\subsubsection{Evaluation for Bag-Track}
Similar to sent-track, given a prediction result at bag-level on test data, let $N_{sys}$ be the number of non-NA relation labels in all bags of prediction results, $N_{std}$ be the number of non-NA relation labels in all bags of standard answer, $N_r$ be the number of relation lables that be correctly predicted.
Finally, the evaluation metrics of bag-track is formally the same as equation (\ref{evalution_metrics}).

\subsection{Leaderboard Strategy}
In this shared task on inter-personal relationship extraction, both sent-track and bag-track have public leaderboard and private leaderboard. 
Specifically, we release all the test data from the beginning of task.
For the test data, we divide it into two parts by about 50\% and keep the distribution of data as consistent as possible.
How to divide test data is also not public, and F1-score will be calculated on fixed 50\% data and 100\% data.
For each prediction results submitted by contestants, there are two F1-scores dubbed as A-score and B-score, in which A-score is calculated on fixed half test data while B-score is on the whole data.
When calculating A-score, prediction results of the other half test data is ignored.
The public leaderboard shows the highest A-scores of all teams in descending order, while the private leaderboard is ranked by B-scores corresponding to A-scores.
After the end of the competition, we will announce the private leaderboard, which is the final ranking.

It is worth mentioning that instances used to calculate A-score in sent-track and bag-track are identical, which effectively avoids the risk of leaking answers.
In addition, there are amounts of automatically constructed NA data in the test set, which basically eliminates the possibility of annotating test data manually.

\section{Methodology}
\label{section_approaches}
The shared task provides a baseline system \footnote{\url{https://github.com/ccks2019-ipre/baseline}} for both sent-track and bag-track based on convolutional neural network (CNN).
In this section, we give a brief overview to the approaches used by the participating teams, and finally give the results of the top 10 and baseline of the task.

\subsection{Main Technologies}

\subsubsection{BERT}
Since the release of BERT \cite{devlin2018bert}, it has quickly surpassed ELMo \cite{peters2018deep}, becoming a hot language model in the NLP field, and has achieved good performance in multiple NLP tasks.
Compared with the traditional methods for training word vector, such as Skip-Gram \cite{mikolov2013efficient} and GloVe \cite{pennington2014glove}, BERT can provide richer and stronger contextual representation. 
In the shared task, it yields relatively high performances that simply add a dense layer after BERT to classify or use the output of BERT as word embeddings. 
Considering that BERT may overfit the entity names, Team LEKG replaces the given entity pair in each sentence with two fixed names, ``刘伟明" (Liu Weiming) and ``李静平" (Li Jingping).

\subsubsection{Track Conversion}
Since sent-track and bag-track share the same datasets, and relation extraction at bag-level shows a better noise immunity, many teams convert the results of bag-track into the results into sent-track.
The key of conversion is to check whether the sentences in the bag express the relation type of bag or not.
Training a binary classifier is an intelligent choice, and using feature templates is also a good choice.
Team NEU\_DM1 and Team ``格物致知" (GWZZ) use these methods respectively and their experiments show that their motheds can also achieve good results.

\subsubsection{Feature Engineering}
Feature engineering is an effective way to improve the performance of system, requiring careful analysis of raw data.
Participating teams have provided many good feature-based methods to inspire us.
In terms of inter-personal relationship, some features, such as gender or surname, are easily utilized to predict the relationship between persons due to the nature of data.
There is an open source tool \footnote{\url{https://pypi.org/project/ngender}} that can judge gender based on Chinese names and it perfroms well.
Surname helps to judge blood relationship, especially the relationship between father and son.
Correcting the answer with limited keywords also achieves very good results.
We can initially predict the relationship between persons through these keywords such as ``夫" (husband), ``妻" (wife) and so on.

\begin{table}[tb]
	\centering
	\footnotesize
	\caption{Top 10 in sent-track and bag-track}
	\label{ranking}
	\begin{tabular}{|c|c|c|c|c|c|c|c|}
		\hline
		\multicolumn{4}{|c|}{Sent-Track} & \multicolumn{4}{c|}{Bag-Track} \\ \hline
		\ Ranking\  & Team &\ A-score\  &\  B-score\  &\  Ranking\  & Team &\  A-score\  &\  B-score \ \\ \hline
		\# 1 & 格物致知 & 0.54076 & 0.54279 & \# 1 & LEKG & 0.59925 & 0.63030 \\ \hline
		\# 2 & LEKG & 0.47300 & 0.48427 & \# 2 & 格物致知 & 0.60773 & 0.62162 \\ \hline
		\# 3 & NEU\_DM1 & 0.44912 & 0.46200 & \# 3 & NEU\_DM1 & 0.55894 & 0.57459 \\ \hline
		\# 4 & LMN & 0.41171 & 0.41096 & \# 4 & Ac & 0.51899 & 0.53196 \\ \hline
		\# 5 & RE小分队 & 0.42841 & 0.41003 & \# 5 & idke\_NEU & 0.49724 & 0.52374 \\ \hline
		\# 6 & runit & 0.40297 & 0.39523 & \# 6 & Jun & 0.50785 & 0.52346 \\ \hline
		\# 7 & Jun & 0.38566 & 0.38657 & \# 7 & OneOf & 0.48895 & 0.50165 \\ \hline
		\# 8 & guanchong & 0.40322 & 0.38044 & \# 8 & jack & 0.49351 & 0.49038 \\ \hline
		\# 9 & 机器没有命运 & 0.37741 & 0.35767 & \# 9 & guanchong & 0.45783 & 0.47665 \\ \hline
		\# 10 & uw1 & 0.35834 & 0.34885 & \# 10 & 华凌NLP & 0.44852 & 0.47612 \\ \hline
		\# 78 & baseline & 0.20564 & 0.20539 & \# 30 & basline & 0.33237 & 0.33090 \\ \hline
	\end{tabular}
\end{table}

\subsubsection{Data Preprocessing}
In consideration of the unbalanced distribution of relations, some teams such as Team NEU\_DM1 delete nearly half of the relations in order to reduce the impact of these relations, which perfroms well.
Moreover, in order to enrich courpus, Team ``RE小分队" (RE) employs translator tools \footnote{\url{https://fanyi.baidu.com/}} to convert it into other languages and translate it back to Chinese.
The method of undersampling and oversampling can also alleviate the problem of the imbalance of data, which is utilized by Team GWZZ.


\begin{table}[tph]
	\centering
	\footnotesize
	\caption{Results of top 3 in sent-track}
	\label{tab:top3-sent-track}
	\begin{tabular}{|c|c|c|c|c|c|c|c|}
		\hline
		\multirow{2}{*}{RelationID} & \multicolumn{1}{c|}{\multirow{2}{*}{Test (Sent-Track)}} & \multicolumn{2}{c|}{Top 1} & \multicolumn{2}{c|}{Top 2} & \multicolumn{2}{c|}{Top 3} \\ \cline{3-8} 
		& \multicolumn{1}{c|}{} & \  Sentence \  &\  F1\  &\   Sentence\   &\ F1 \  &\   Sentence\   & \  F1\  \\ \hline
		\textbf{\# 1} & \textbf{536} & \textbf{649} & \textbf{0.69030} & \textbf{695} & \textbf{0.60601} & \textbf{701} & \textbf{0.55133}
		\\ \hline
		\# 2 & 12 & 9 & 0.76190 & 0 & 0.00000 & 0 & 0.00000 \\ \hline
		\# 3 & 13 & 0 & 0.00000 & 0 & 0.00000 & 0 & 0.00000 \\ \hline
		\textbf{\# 4} & \textbf{296} & \textbf{383} & \textbf{0.57732} & \textbf{333} & \textbf{0.56916} & \textbf{307} & \textbf{0.50746} \\ \hline
		\# 5 & 3 & 4 & 0.57143 & 0 & 0.00000 & 0 & 0.00000 \\ \hline
		\# 6 & 10 & 0 & 0.00000 & 0 & 0.00000 & 0 & 0.00000 \\ \hline
		\# 7 & 9 & 41 & 0.20000 & 2 & 0.00000 & 63 & 0.13889 \\ \hline
		\# 8 & 1 & 0 & 0.00000 & 0 & 0.00000 & 0 & 0.00000 \\ \hline
		\# 9 & 0 & 0 & 0.00000 & 0 & 0.00000 & 0 & 0.00000 \\ \hline
		\textbf{\# 10} & \textbf{490} & \textbf{1,254} & \textbf{0.44266} & \textbf{1,226} & \textbf{0.41026} & \textbf{1,172} & \textbf{0.39832} \\ \hline
		\textbf{\# 11} & \textbf{107} & \textbf{152} & \textbf{0.64865} & \textbf{122} & \textbf{0.54148} & \textbf{161} & \textbf{0.59701} \\ \hline
		\textbf{\# 12} & \textbf{221} & \textbf{270} & \textbf{0.58248} & \textbf{212} & \textbf{0.54503} & \textbf{292} & \textbf{0.55361} \\ \hline
		\# 13 & 63 & 119 & 0.63736 & 34 & 0.43299 & 101 & 0.64634 \\ \hline
		\# 14 & 4 & 3 & 0.57143 & 0 & 0.00000 & 0 & 0.00000 \\ \hline
		\# 15 & 8 & 2 & 0.20000 & 0 & 0.00000 & 0 & 0.00000 \\ \hline
		\# 16 & 43 & 90 & 0.51128 & 72 & 0.31304 & 98 & 0.29787 \\ \hline
		\# 17 & 39 & 52 & 0.54945 & 15 & 0.48148 & 42 & 0.37037 \\ \hline
		\# 18 & 9 & 15 & 0.66667 & 0 & 0.00000 & 6 & 0.26667 \\ \hline
		\# 19 & 23 & 64 & 0.32184 & 22 & 0.08889 & 62 & 0.32941 \\ \hline
		\# 20 & 1 & 0 & 0.00000 & 0 & 0.00000 & 0 & 0.00000 \\ \hline
		\# 21 & 0 & 1 & 0.00000 & 0 & 0.00000 & 0 & 0.00000 \\ \hline
		\# 22 & 2 & 0 & 0.00000 & 0 & 0.00000 & 0 & 0.00000 \\ \hline
		\# 23 & 0 & 0 & 0.00000 & 0 & 0.00000 & 0 & 0.00000 \\ \hline
		\# 24 & 0 & 0 & 0.00000 & 0 & 0.00000 & 0 & 0.00000 \\ \hline
		\# 25 & 0 & 0 & 0.00000 & 0 & 0.00000 & 0 & 0.00000 \\ \hline
		\# 26 & 4 & 0 & 0.00000 & 0 & 0.00000 & 0 & 0.00000 \\ \hline
		\# 27 & 0 & 0 & 0.00000 & 0 & 0.00000 & 0 & 0.00000 \\ \hline
		\# 28 & 0 & 0 & 0.00000 & 0 & 0.00000 & 0 & 0.00000 \\ \hline
		\# 29 & 1 & 0 & 0.00000 & 0 & 0.00000 & 0 & 0.00000 \\ \hline
		\# 30 & 27 & 1 & 0.07143 & 4 & 0.25806 & 1 & 0.00000 \\ \hline
		\# 31 & 44 & 0 & 0.00000 & 0 & 0.00000 & 23 & 0.25806 \\ \hline
		\# \textbf{32} & \textbf{149} & \textbf{28} & \textbf{0.13559} & \textbf{0} & \textbf{0.00000} & \textbf{125} & \textbf{0.14599} \\ \hline
		\textbf{\# 33} & \textbf{201} & \textbf{191} & \textbf{0.73469} & \textbf{198} & \textbf{0.67669} & \textbf{165} & \textbf{0.68852} \\ \hline
		\# 34 & 11 & 1 & 0.52174 & 16 & 0.07407 & 38 & 0.20408 \\ \hline
	\end{tabular}
\end{table}

\begin{table}[tph]
	\centering
	\footnotesize
	\caption{Results of top 3 in bag-track}
	\label{tab:top3-bag-track}
	\begin{tabular}{|c|c|c|c|c|c|c|c|}
		\hline
		\multirow{2}{*}{RelationID} & \multicolumn{1}{c|}{\multirow{2}{*}{\ Test (Bag-Track)\ }} & \multicolumn{2}{c|}{Top 1} & \multicolumn{2}{c|}{Top 2} & \multicolumn{2}{c|}{Top 3} \\ \cline{3-8} 
		& \multicolumn{1}{c|}{} & \ \  Triple \ \  &\ \  F1\ \  &\ \   Triple\ \   &\ \ F1 \ \  &\  \  Triple\  \  & \ \  F1\ \  \\ \hline
		\textbf{\# 1} & \textbf{112} & \textbf{140} & \textbf{0.70635} & \textbf{106} & \textbf{0.77982} & \textbf{112} & \textbf{0.73214}
		\\ \hline
		\# 2 & 7 & 0 & 0.00000 & 4 & 0.54545 & 0 & 0.00000 \\ \hline
		\# 3 & 11 & 0 & 0.00000 & 0 & 0.00000 & 0 & 0.00000 \\ \hline
		\textbf{\# 4} & \textbf{71} & \textbf{82} & \textbf{0.67974} & \textbf{63} & \textbf{0.64179} & \textbf{79} & \textbf{0.66667} \\ \hline
		\# 5 & 3 & 0 & 0.00000 & 2 & 0.80000 & 0 & 0.00000 \\ \hline
		\# 6 & 9 & 0 & 0.00000 & 0 & 0.00000 & 0 & 0.00000 \\ \hline
		\# 7 & 7 & 14 & 0.28571 & 14 & 0.28571 & 21 & 0.21429 \\ \hline
		\# 8 & 1 & 0 & 0.00000 & 0 & 0.00000 & 0 & 0.00000 \\ \hline
		\# 9 & 0 & 0 & 0.00000 & 0 & 0.00000 & 0 & 0.00000 \\ \hline
		\textbf{\# 10} & \textbf{154} & \textbf{250} & \textbf{0.70297} & \textbf{210} & \textbf{0.64835} & \textbf{203} & \textbf{0.64426} \\ \hline
		\textbf{\# 11} & \textbf{39} & \textbf{49} & \textbf{0.70455} & \textbf{43} & \textbf{0.75610} & \textbf{37} & \textbf{0.50000} \\ \hline
		\textbf{\# 12} & \textbf{51} & \textbf{81} & \textbf{0.72727} & \textbf{61} & \textbf{0.625} & \textbf{57} & \textbf{0.62963} \\ \hline
		\# 13 & 28 & 31 & 0.81356 & 30 & 0.82759 & 31 & 0.71186 \\ \hline
		\# 14 & 2 & 1 & 0.00000 & 1 & 0.66667 & 0 & 0.00000 \\ \hline
		\# 15 & 2 & 0 & 0.00000 & 1 & 0.66667 & 0 & 0.00000 \\ \hline
		\# 16 & 24 & 43 & 0.53731 & 29 & 0.64151 & 42 & 0.42424 \\ \hline
		\# 17 & 14 & 17 & 0.58065 & 12 & 0.69231 & 18 & 0.4375 \\ \hline
		\# 18 & 6 & 6 & 0.66667 & 7 & 0.76923 & 13 & 0.52632 \\ \hline
		\# 19 & 11 & 13 & 0.50000 & 14 & 0.56000 & 14 & 0.48000 \\ \hline
		\# 20 & 1 & 0 & 0.00000 & 0 & 0.00000 & 0 & 0.00000 \\ \hline
		\# 21 & 0 & 0 & 0.00000 & 1 & 0.00000 & 0 & 0.00000 \\ \hline
		\# 22 & 2 & 0 & 0.00000 & 0 & 0.00000 & 0 & 0.00000 \\ \hline
		\# 23 & 0 & 0 & 0.00000 & 0 & 0.00000 & 0 & 0.00000 \\ \hline
		\# 24 & 0 & 0 & 0.00000 & 0 & 0.00000 & 0 & 0.00000 \\ \hline
		\# 25 & 0 & 0 & 0.00000 & 0 & 0.00000 & 0 & 0.00000 \\ \hline
		\# 26 & 2 & 0 & 0.00000 & 0 & 0.00000 & 0 & 0.00000 \\ \hline
		\# 27 & 0 & 0 & 0.00000 & 0 & 0.00000 & 0 & 0.00000 \\ \hline
		\# 28 & 0 & 0 & 0.00000 & 0 & 0.00000 & 0 & 0.00000 \\ \hline
		\# 29 & 1 & 0 & 0.00000 & 0 & 0.00000 & 0 & 0.00000 \\ \hline
		\# 30 & 12 & 5 & 0.35294 & 1 & 0.15385 & 5 & 0.00000 \\ \hline
		\# 31 & 29 & 3 & 0.12500 & 0 & 0.00000 & 1 & 0.06667 \\ \hline
		\textbf{\# 32} & \textbf{72} & \textbf{22} & \textbf{0.23404} & \textbf{11} & \textbf{0.16867} & \textbf{18} & \textbf{0.26667} \\ \hline
		\textbf{\# 33} & \textbf{61} & \textbf{54} & \textbf{0.76522} & \textbf{47} & \textbf{0.74074} & \textbf{51} & \textbf{0.78571} \\ \hline
		\# 34 & 8 & 6 & 0.71429 & 9 & 0.58824 & 6 & 0.42857 \\ \hline
	\end{tabular}
\end{table}

\subsection{Results and Analysis}
Due to the space limitation, we only list the scores of top 10 in sent-track and bag-track.
As shown in Table \ref{ranking}, approaches used by the participating teams performs much better than baseline.
It is worth noting that there are some changes in the final ranking due to our evaluation strategy.

Specifically, we analyze the results of the top 3 in sent-track and bag-track, and detailed information are listed in Table \ref{tab:top3-sent-track} and Table \ref{tab:top3-bag-track} respectively.
As mentioned above, whether predicted results have a NA label or not will not affect the calcultion of F1-score, and we can also see the unbalanced distribution of relations from the Table \ref{relations}.
Hence, if a predicted result has good performance in some large categories, it will have a good effect on the whole.
Seven relations account for most of the test data, they are \emph{present husband} (\# 1), \emph{present wife} (\# 4), \emph{father} (\# 10), \emph{mother} (\# 11), \emph{son} (\#12), \emph{lover} (\# 32) and \emph{teacher} (\# 33).
In sent-track, 2,000 of the 2,300 standard answers belong to these seven relations.
In bag-track, these seven relations contain 560 of the 740 standard answers.

As shown in Table \ref{tab:top3-sent-track} and Table \ref{tab:top3-bag-track}, top 3 give up the prediction of some small categories and concentrate on the prediction of large categories, especially these seven relations.
We compare the performance of each system in different relations by F1-score.
All systems show good results in most of these seven relations. 
It can be seen that the predicted results of these reations are positively correlated with the overall results. 
One difference is that all three systems perform poorly in the prediction of the relation \emph{lover}.
We randomly sample some test sentences that actually express the relation of \emph{lover}.
Although there are some decisive keywords in these sentences, such as ``相恋" (fall in love), ``男友" (boyfriend), ``女友" (girlfriend), these systems tend to predict them as NA rather than \emph{lover}.
By contrast, systems prefer to predict NA as \emph{father}, and the number of sentences or triples which predicted as father exceeds the actual number of ones that express the relation of \emph{father} in both sent-track and bag-track.
A manual check indicates that large amounts of sentences which just express the relation between parent and children are misclassified as the relation \emph{father}. 
False negatives also have some influence on this result.

\begin{figure}[tb]
	\centering
	\includegraphics[width=0.72\textwidth]{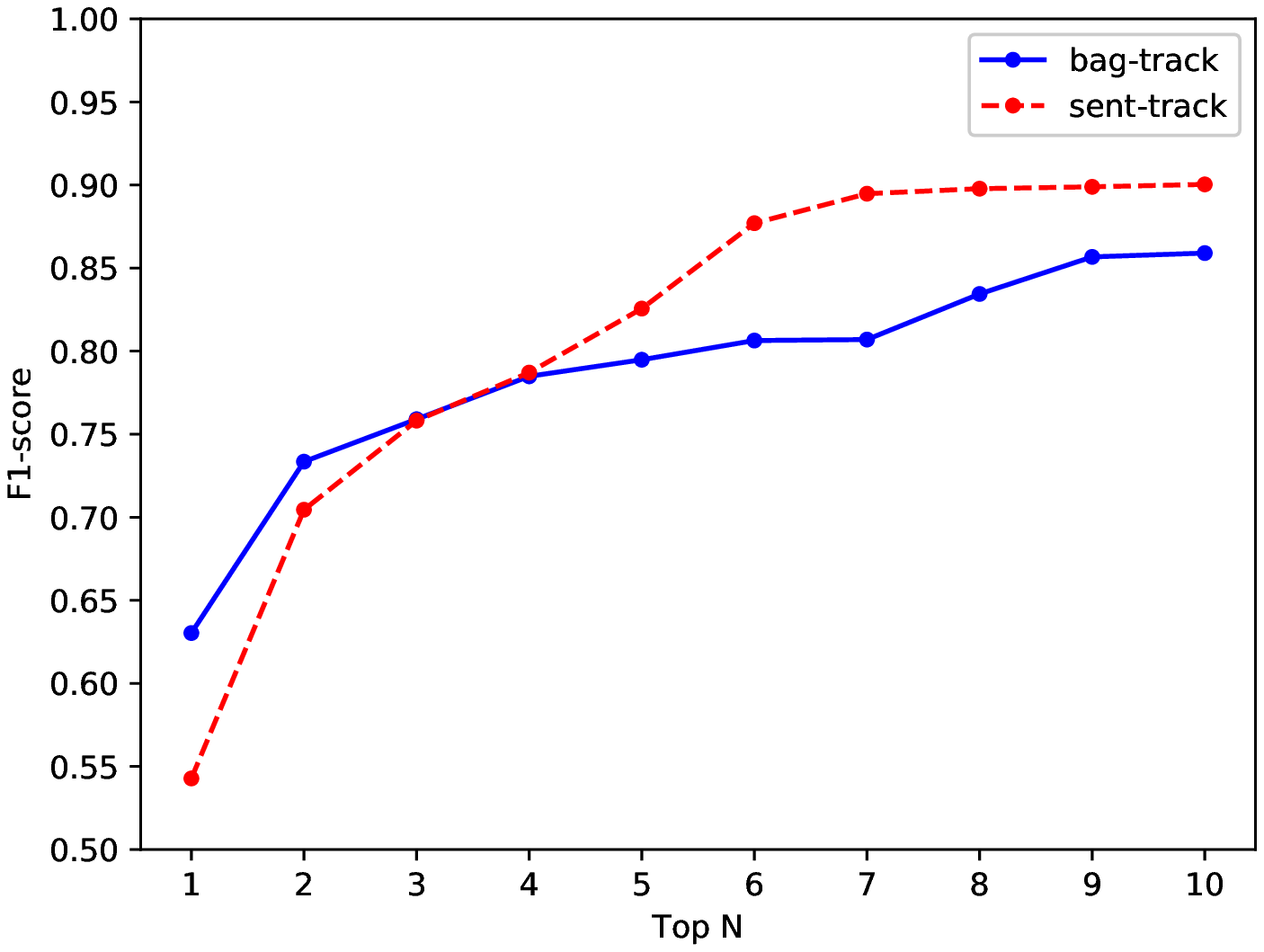}
	\caption{Upper bounds of top N systems.} 
	\label{fig:upper_bounds}
\end{figure}

In order to provide a reference upper bound for future research on the shared task, we combine the results of the top 10 participating systems both on sent-track and bag-track.
For each entry, If at least one of top N systems gives out the correct answer, we count it as correct.
As shown in Figure \ref{fig:upper_bounds}, we can see that with the number of fusion results increases, the value of F1 increases dramatically.
It is surprising that the upper bound of the sent-track finally exceeds the bag-track, which may be caused by the fact that a bag tends to have more than one relations.
Compared to a sentence or bag with only one relation, a bag with multiple relations is more difficult to predict all relations.

\section{Conclusion}
\label{section_conclusion}
The CCKS 2019 Shared Task on Inter-Personal Relationship Extraction consists of sent-track and bag-track.
This year, 358 teams from various universities and organizations participated in the task.
The goal of the task is to predict the relationship between two person entitties through the given sentence.
The best system achieves an F1 score of 54.3\% in the sent-track, and 63.0\% in the bag-track. 
The IPRE dataset used in this shared task can be found at \url{https://github.com/SUDA-HLT/IPRE}. 
We refer to the system description papers for more in-depth analysis of individual systems and their performance in details \cite{liu2019fmpk,peng2019abert,shen2019bert,zhu2019improving}.
\section{Acknowledgements}
The research work is supported by the National Natural Science Foundation of China (Grant No. 61525205, 61876115). 
China Conference on Knowledge Graph and Semantic Computing provides a forum for the shared task and Biendata \footnote{\url{https://biendata.com}} also provides technical support.  
We would like to thank all the contributors to Inter-Personal Relationship Extraction, including all participating teams and workers of data annotation, without their effort a task like this simply wouldn’t be possible.

%
%
%
 \bibliographystyle{splncs04}
 \bibliography{ipre}
%
%
%
%
%
\end{CJK}
\end{document}